\documentclass[conference]{IEEEtran}
\usepackage{amsmath,amssymb,amsfonts}
\usepackage{algorithmic}
\usepackage{graphicx}
\usepackage{textcomp}
\usepackage{xcolor}
\usepackage{tikz}
\usepackage{pgfplots}
\usepackage{cite}
\usepackage{url}
\usepackage{siunitx}
\usepackage{tabularx}
\usepackage{array} 
\newcolumntype{L}{>{\raggedright\arraybackslash}X}
\usepackage[caption=false,font=footnotesize]{subfig}
\pgfplotsset{compat = 1.11}
\usepackage[export]{adjustbox}
\usepackage{booktabs}

\usetikzlibrary{chains, positioning, fit}
\def\BibTeX{{\rm B\kern-.05em{\sc i\kern-.025em b}\kern-.08em
		T\kern-.1667em\lower.7ex\hbox{E}\kern-.125emX}}
\usepackage{etoolbox}
\usepackage{pbalance}
\makeatother

\usepackage{fancyhdr}
\usepackage[hidelinks]{hyperref}

\fancypagestyle{firstpage}
{
    \fancyhf{} 
    \renewcommand{\footrulewidth}{0.5pt}
\renewcommand{\footrule}{\hbox to \headwidth{\color{red}\leaders\hrule height \footrulewidth\hfill}}
    \fancyfoot[L]{
        \begin{minipage}[t]{0.49\textwidth}
            \scriptsize 
            \noindent 
            \textsf{\textcolor{red}{© 2024 IEEE. Personal use of this material is permitted. Permission from IEEE must be obtained for all other uses, in any current or future media, including printing/republishing this material for advertising or promotional purposes, creating new collective works, for resale or redistribution to servers or lists, or reuse of any copyrighted component of this work in other works.}}
        \end{minipage}
    }
    \fancyfoot[R]{
        \begin{minipage}[t]{0.4825\textwidth}
            \scriptsize 
            \noindent 
            \textsf{\textcolor{red}{This is the author's accepted manuscript. Full citation for the published version is included here: S. Novaković and V. Risojević, "Learning Localization of Body and Finger Animation Skeleton Joints on Three-Dimensional Models of Human Bodies," 2024 Zooming Innovation in Consumer Technologies Conference (ZINC), Novi Sad, Serbia, 2024, pp. 19-24, DOI:} \href{https://ieeexplore.ieee.org/abstract/document/10579426}{\textcolor{red}{10.1109/ZINC61849.2024.10579426}}\textcolor{red}{, URL:} \href{https://ieeexplore.ieee.org/abstract/document/10579426}{\textcolor{blue}{https://ieeexplore.ieee.org/abstract/document/10579426}}}
        \end{minipage}
    }

}

\begin{document}

\title{Learning Localization of Body and Finger Animation Skeleton Joints on Three-Dimensional Models of Human Bodies}

\author{\IEEEauthorblockN{Stefan Novaković}
\IEEEauthorblockA{\textit{Faculty of Electrical Engineering} \\
\textit{University of Banja Luka}\\
Banja Luka, Bosnia and Herzegovina \\
stefan.novakovic@etf.unibl.org}%
\and
\IEEEauthorblockN{Vladimir Risojević}
\IEEEauthorblockA{\textit{Faculty of Electrical Engineering} \\
\textit{University of Banja Luka}\\
Banja Luka, Bosnia and Herzegovina \\
vladimir.risojevic@etf.unibl.org}
}

\maketitle

\thispagestyle{firstpage}

\begin{abstract}
Contemporary approaches to solving various problems that require analyzing three-dimensional (3D) meshes and point clouds have adopted the use of deep learning algorithms that directly process 3D data such as point coordinates, normal vectors and vertex connectivity information. Our work proposes one such solution to the problem of positioning body and finger animation skeleton joints within 3D models of human bodies. Due to scarcity of annotated real human scans, we resort to generating synthetic samples while varying their shape and pose parameters. Similarly to the state-of-the-art approach, our method computes each joint location as a convex combination of input points. Given only a list of point coordinates and normal vector estimates as input, a dynamic graph convolutional neural network is used to predict the coefficients of the convex combinations. By comparing our method with the state-of-the-art, we show that it is possible to achieve significantly better results with a simpler architecture, especially for finger joints. Since our solution requires fewer precomputed features, it also allows for shorter processing times.
\end{abstract}


\section{Introduction}

Three-dimensional (3D) models of human bodies have various use cases in a diverse range of fields: animation, healthcare, medical research, anthropometry, virtual reality, digital avatars, etc. They may be designed using 3D modeling software, generated through parametric models \cite{SMPL:2015, MANO:SIGGRAPHASIA:2017, SMPL-X:2019, STAR:2020, 50649, SUPR:2022, MakeHuman, 9603558}, or derived as a result of 3D scanning \cite{805368, 6682899, 6909880, zchao, 8491001, yan2019anthropometric, Zheng2019DeepHuman, tiwari20sizer, tao2021function4d, Jinka2022SHARPSR, 10125586, 9760157}.

\begin{figure}
	\centering
	\adjincludegraphics[width=0.28\textwidth, trim={{.359375\width} 0 {.27760416666666666666666666666667\width} 0}]{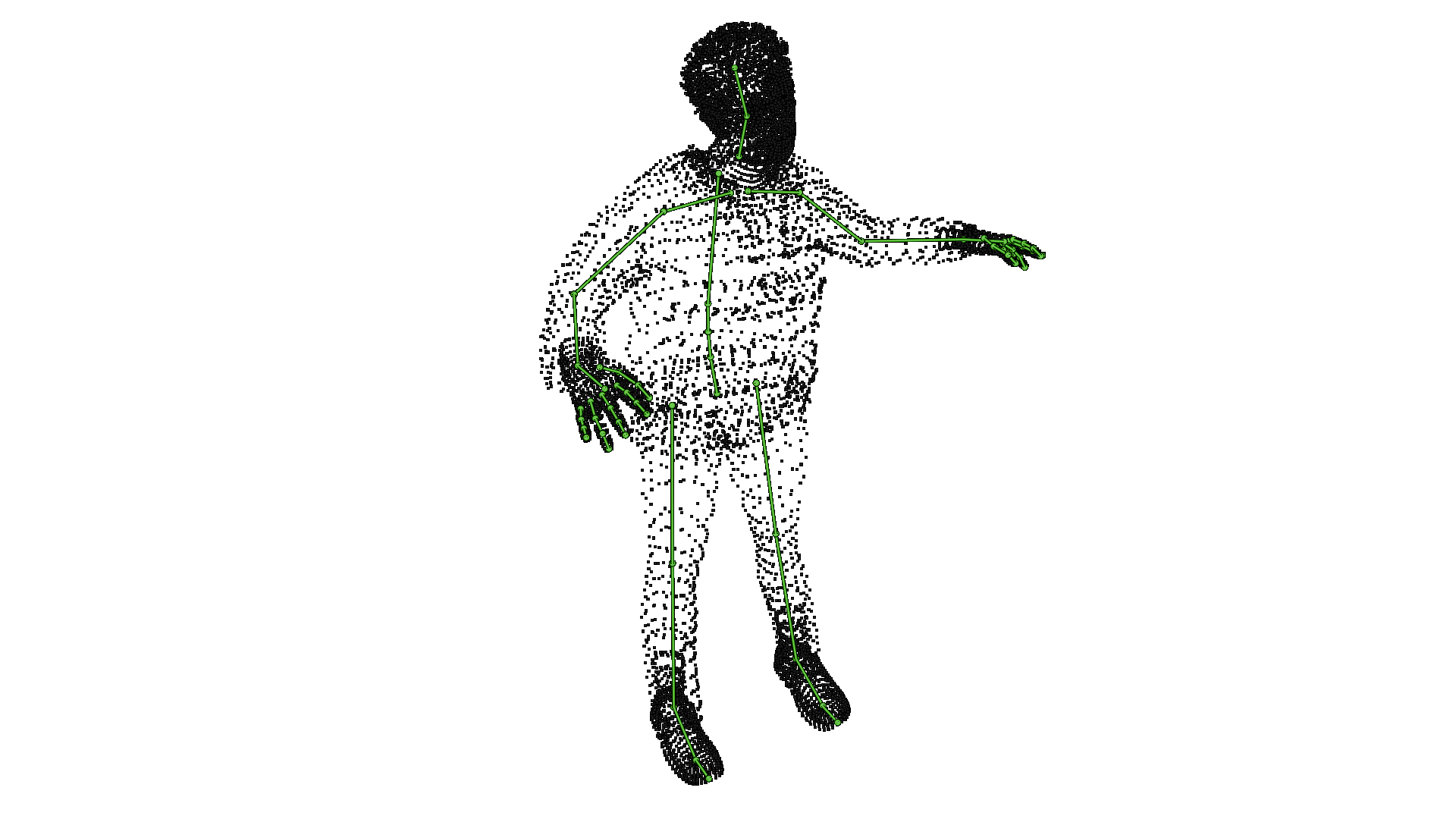}
	\caption{Visualization of input point cloud overlaid with output joint estimations connected into a skeletal hierarchy (colored green), rendered using Blender \cite{Blender} viewport rendering.}
	\label{figure:example}
\end{figure}

In the context of computer animation, the animation skeleton of a 3D model is a compact abstraction which comprises a hierarchy of skeletal joints, with adjacent joints forming the skeletal bones. Categories of animation skeleton joints for human bodies typically include: head, neck, shoulders, spine, hips, elbows, hands, wrists, knees, feet and fingers \cite{ISOIEC19774-1:2019, ISOIEC19774-2:2019, 10.5555/2385444}.

In order to be able to animate a 3D model using an animation skeleton, one of the first steps involves fitting the skeleton to the 3D model in the process known as rigging. Often, such fitted skeleton is an instance of a template skeleton, having a predefined set of joint types. Rigging is performed by placing individual joints at the appropriate locations within the 3D mesh. Manually fitting a template skeleton to a large number of 3D scans of human bodies is a long and tedious process. This has led to the development of solutions that automate parts of this process. Contemporary research on this subject follows two directions: 1) fitting a parametric 3D model of human body to the target 3D scan and reusing the skeleton from the fitted 3D model \cite{bhatnagar2020ipnet, bhatnagar2020loopreg} and 2) end-to-end deep learning of joint localization \cite{RigNet, MA2023158}.

Our main contribution is a neural network architecture for localizing body and finger animation skeleton joints, given a point cloud as input. We also provide a method for generating synthetic 3D models of human bodies with randomized shape and pose parameters. Using such a dataset for training and evaluation purposes, we show that our solution achieves significantly better results when compared to the state-of-the-art. An example input-output pair is shown in Fig.~\ref{figure:example}.

\section{Related work}

\subsection{Point cloud analysis}

PointNet \cite{8099499} is a seminal work on the subject of point cloud analysis, initially used for point cloud classification and segmentation problems. It introduces the application of a shared multilayer perceptron (MLP) for per-point feature extraction, followed by a feature-wise maximum pooling operator. In this way, the point cloud characteristics are condensed into a global feature vector, achieving permutation invariance.

In \cite{dgcnn}, a dynamic graph convolutional neural network (DGCNN) is proposed for point cloud analysis. It introduces the EdgeConv module, a graph convolution operator intended for use on point sets. The point set needs to be reformulated as a graph, with its vertices being the points and its edges forming neighbourhoods of points. Given a set of point feature vectors, the $k$-nearest neighbours algorithm can be used to compute, for each point, the edges of the graph. Inspired by PointNet, EdgeConv computes each point feature by applying a shared MLP over the edge features of the point, followed by maximum pooling. At each EdgeConv block, the neighbourhoods are recomputed based on distance in feature space. Chaining multiple EdgeConv blocks in a sequence facilitates grouping of points with similar embeddings. The initial neighbourhoods are often determined by Euclidean distance in 3D space. The published code libraries of PointNeXt \cite{qian2022pointnext} and PointMeta \cite{10205335} contain a DGCNN implementation that we use in our work.

In \cite{lt3dhpfpc}, the problem of 3D human pose estimation from depth images is tackled. They propose estimating 3D point clouds from the 2D depth images and directly regressing the 3D joint positions with a DGCNN-based backbone. However, in contrast to our problem, their approach assumes partial and noisy point clouds, considering they are reconstructed from a limited amount of viewpoints.

\subsection{Neural rigging}

RigNet \cite{RigNet} introduces a DGCNN-based non-template skeletal joint prediction module that employs an iterative variant of mean-shift clustering \cite{400568} to collapse mesh vertices into joint locations. The method relies on a combination of point displacement learning \cite{yin2018p2pnet}, attention pretraining and surface geodesic distance computation, which they implement as a $\Theta(n^2\log{n})$ operation. By design, this method is not suitable for template joint localization.

In the TARig architecture \cite{MA2023158}, the template joint module repurposes the RigNet DGCNN-based joint prediction backbone for template joint prediction, discarding the iterative mean-shift clustering approach and localizing the template joints as convex combinations of the mesh vertices.

Both RigNet and TARig focus on a variety of shapes emerging from a dataset of stylized characters in symmetric poses (e.g. T-pose), achieving state-of-the-art results. They do not specifically train or evaluate the model for realistic characters, nor do they attempt to consider slight pose variations, which is something that may occur in 3D scans of people. In their work, they also do not consider prediction of finger joints that require finer estimation compared to body joints.

\section{Architecture}
\label{sec:arch}
\begin{figure*}[!t]
	\centering
	\begin{tikzpicture}[node distance=1cm]
  		\tikzstyle{node}=[draw, minimum size=1cm, minimum width=2cm, align=center]

		\begin{scope}[start chain=going right, node distance=1cm]
			\node[on chain] (n0) at (0, 0) {\shortstack{Points \\ $N \times 3$}};

			\node[minimum size=2cm, inner sep=0pt, on chain, outer sep=0pt, draw=none, right=0.25cm of n0] (cat0carrier) {};
			\node[draw, circle, minimum size=1cm, minimum width=1 cm,  inner sep=0pt, outer sep=0pt] at (cat0carrier.center) (cat0) {};
			\draw (cat0.west) -- (cat0.east);
			\draw (cat0.north) -- (cat0.south);

			\node[on chain, above=0.5cm of cat0] (nn) {\shortstack{Normals \\ $N \times 3$}};

			\draw[-stealth] (n0) -- (cat0) node[midway, above] {};
			\draw[-stealth] (nn) -- (cat0) node[midway, above] {};

			\node[node, on chain, right=0.725cm of cat0] (ec1) {EdgeConv\\$(6, 64)$};
			\node[node, on chain] (ec2) {EdgeConv\\$(64, 64)$};
			\node[node, on chain] (ec3) {EdgeConv\\$(64, 128)$};
			\node[node, on chain] (ec4) {EdgeConv\\$(128, 256)$};

			\draw[-stealth] (cat0) -- (ec1) node[midway, above] {\tiny{$N \times 6$}};
			\draw[-stealth] (ec1) -- (ec2) node[midway, above] {\tiny{$N \times 64$}};
			\draw[-stealth] (ec2) -- (ec3) node[midway, above] {\tiny{$N \times 64$}};
			\draw[-stealth] (ec3) -- (ec4) node[midway, above] {\tiny{$N \times 128$}};
		\end{scope}

		\node[fit=(ec1)(ec4), inner sep=0pt] (fit1) {};

		\node[minimum size=2cm, inner sep=0pt, on chain, outer sep=0pt, draw=none, right=1cm of fit1] (catcarrier) {};

		\node[draw, circle, minimum size=1cm, minimum width=1 cm,  inner sep=0pt, outer sep=0pt] at (catcarrier.center) (cat) {};

		\draw (cat.west) -- (cat.east);
		\draw (cat.north) -- (cat.south);

       	\foreach \i in {1,...,3}
			\draw[-stealth] (ec\i.north) -- ++(0,+0.5cm) -| (cat);
		\draw[-stealth] (ec4) -- (cat) node[midway, above] {\tiny{$N \times 256$}};

		\node[fit=(fit1)(cat), inner sep=0pt] (fit2) {};

		\begin{scope}[start chain=going left, node distance=1cm]
			\node[node, on chain, below=0.5cm of catcarrier] (smlp) {SharedMLP\\$(512, 512)$};

			\draw[-stealth] (cat) -- (smlp) node[midway, left] {\tiny{$N \times 512$}};

			\node[node, on chain] (lin) {Linear\\$(512, J)$};

			\draw[-stealth] (smlp) -- (lin) node[midway, above] {\tiny{$N \times 512$}};

			\node[node, on chain] (softmax) {Per-joint\\softmax};

			\draw[-stealth] (lin) -- (softmax) node[midway, above] {\tiny{$N \times J$}};

			\node[node, on chain] (matmul) {Matrix\\multiply};

			\draw[-stealth] (softmax) -- (matmul) node[midway, above] {\tiny{$N \times J$}};

			\draw[-stealth] (n0) |- ([yshift=0.5cm]matmul.north) -- (matmul.north);

			\node[left=1cm of matmul] (j) {\shortstack{\textbf{Joints} \\ $\mathbf{J \times 3}$}};

			\draw[-stealth] (matmul) -- (j);

		\end{scope}

	\end{tikzpicture}
	\caption{Our proposed architecture based on the DGCNN backbone as provided in the published code of PointNeXt \cite{qian2022pointnext}. As input for the DGCNN-based neural network, we provide the 3D point coordinates and their estimated normal vectors. The first EdgeConv layer computes edges based on $k$-nearest neighbour distances between points in 3D space. Subsequent EdgeConv layers compute edges based on distances in feature space. The output is a list of estimated joint positions. For each joint, its coordinates are computed as a convex combination of the input points. We employ a per-joint softmax layer to guarantee that, for each joint, the convex combination coefficients are non-negative and sum to $1$. }
	\label{figure:architecture}
\end{figure*}
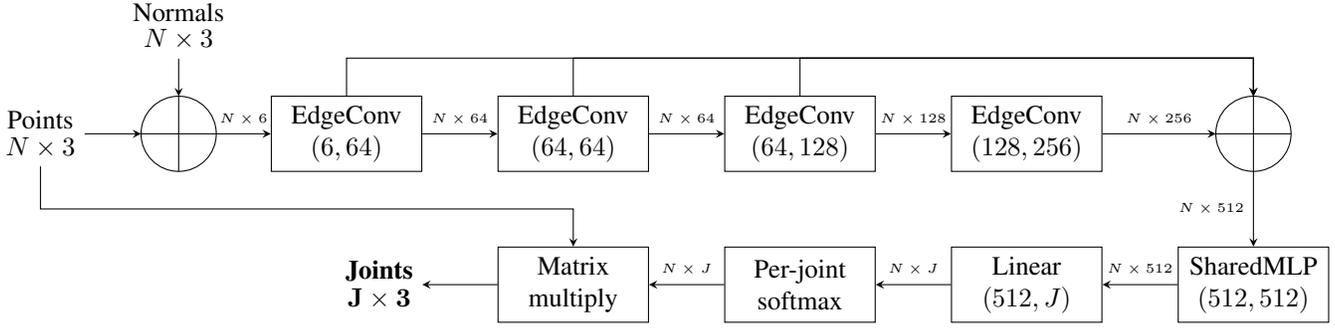

Similarly to the TARig template joint module, our architecture, depicted in Fig.~\ref{figure:architecture}, computes each joint location as a convex combination of the input point positions. This assumes that the joints must be inside the convex hull of the mesh, considering that the convex hull of a point set $S$ is the set of all convex combinations of points in $S$ \cite{Rockafellar+1970}. The convex combination coefficients are estimated by our network.

The proposed architecture employs a basic DGCNN backbone that extracts per-point features, given input point positions and per-point normal vectors. As explained further, the backbone is followed by a per-joint softmax layer used to predict, for each joint, the convex combination coefficients.

Let $J$ be the number of joints. Given $N$ points, the softmax layer outputs an $N \times J$ matrix $A$ in which the $k$-th column contains the convex combination coefficients for the $k$-th joint, $k\in\{1, \ldots, J\}$. The position of the $k$-th joint is computed as
\begin{equation}
	\widehat{\mathbf{j}_k}=\sum_{i=1}^N A_{i,k}\mathbf{p}_i\text{,}
	\label{eq:ccformula}
\end{equation}
where $\{\mathbf{p}_1, \mathbf{p}_2, \ldots, \mathbf{p}_N\}$ is the input point set. The softmax operator guarantees the validity of the constraint
\begin{equation}
	A_{i,k} \geq 0 \quad \land \quad \sum_{i=1}^N A_{i,k}=1\text{,}
\end{equation}
confining the joint inside the input point set convex hull.

The loss function is chosen to be the sum of squared Euclidean distances between groundtruth and predicted joints,
\begin{equation}
	L=\sum_{k=1}^J\lVert{\mathbf{j}_k} - \widehat{\mathbf{j}_k}\rVert_2^2\text{,}
\end{equation}
where $\mathbf{j}_k$ is the position of the $k$-th groundtruth joint.

\section{Dataset}

\subsection{Synthetic data}
Due to the scarcity of available open datasets and proprietary nature of commercial datasets consisting of quality 3D human scans with annotated animation skeletons that contain both body and finger joints, we employ an approach of generating a synthetic dataset. We use the Mass Produce module of MakeHuman \cite{MakeHuman} software to generate a dataset of 3000 synthetic 3D human models. 2400 3D models were used for training, 300 for validation and 300 for testing. During the synthetic 3D model generation process, shape parameters are randomized using a Gaussian distribution as implemented in the Mass Produce module. These parameters include weight, height, muscle mass, as well as parameters for individual body parts, such as head, hips, torso, arms, legs, hands, fingers, etc. Albeit limited in diversity, clothing items are also varied. All generated models are equipped with a 69-joint skeletal rig (Game engine preset), which carries the groundtruth joint location data. It should be noted that automatically generated rigs are, unlike manually created rigs, more consistent in joint placement with respect to the mesh surface, when considering multiple 3D models. This obviates the need for factoring in tolerances due to human error, making them useful when comparing machine learning models for automatic joint placement.

\subsection{Leaf bone length correction}
The generated dataset contains joints located outside of the mesh, mostly affecting leaf bones (head, feet, fingers) and spine points. Since the predicted joints must be inside the convex hull of the mesh, as explained in Section \ref{sec:arch}, we make sure that all groundtruth joints are within the mesh using a technique based on raycasting via the Blender \cite{Blender} scripting API. Specifically, we emit rays originating from the leaf bone parent joints and reduce corresponding leaf bone lengths if the corresponding vectors from ray origin to ray-mesh intersections are shorter than the leaf bone. We set the leaf bone length to $95\%$ of the length of the ray origin to ray-mesh intersection vector. This does not affect the validity of the skeletal rig, since the bone direction and origin are preserved, while the length can be easily adjusted according to specific requirements during postprocessing.

\begin{table*}
	\renewcommand{\arraystretch}{1.3}
	\centering
	\caption{Comparison of methods by MPJPE on the test set (in \% of 3D model height).}
	\label{table:results_mae}
	\begin{tabularx}{\textwidth}{
			l
			L
			L
			L
			L
			L
			L
			L
			L
			L
			L
			L
		}
		\hline
		Method & {Body} & {Fingers} & {Head} & {Neck} & {Shoulder} & {Spine} & {Hips} & {Elbow} & {Wrist} & {Knee} & {Foot} \\
		\hline
		TARig--TJM          & 1.47 & 0.93 & 1.92 & 1.76 & 1.61 & 1.75 & 1.63 & 1.66 & 0.92 & 1.87 & 0.97 \\
		Ours (no normals)   & 0.77 & 0.31 & 1.12 & 0.81 & 0.72 & 1.07 & 0.78 & 0.66 & 0.47 & 0.90 & 0.62 \\
		Ours (with normals) & {\bfseries 0.56} & {\bfseries 0.19} & {\bfseries 0.84} & {\bfseries 0.54} & {\bfseries 0.53} & {\bfseries 0.93} & {\bfseries 0.59} & {\bfseries 0.45} & {\bfseries 0.22} & {\bfseries 0.62} & {\bfseries 0.43} \\
		\hline
	\end{tabularx}
\end{table*}

\begin{figure*}
	\centering
	\begin{tikzpicture}
		\begin{axis}[
			title={Body joints},
			xlabel={Distance threshold factor},
			ylabel={Percentage of Correct Joints},
			width=0.38\textwidth,
			every axis plot/.append style={very thick},
			grid style=dashed,
			ytick={0, 0.1, 0.2, 0.3, 0.4, 0.5, 0.6, 0.7, 0.8, 0.9, 1.0},
			yticklabel={\pgfmathparse{\tick*100}\pgfmathprintnumber{\pgfmathresult}\%},
			ymax=1.05,
			ymajorgrids=true,
			xmajorgrids=true,
			every tick label/.append style={font=\footnotesize},
			]
			\addplot+[red, mark=None] table[x=factor,y=tarigbody, col sep=comma] {iou2.csv};
			\addplot+[orange, mark=None] table[x=factor,y=sdgcnnbodynn, col sep=comma] {iou2.csv};
			\addplot+[green, mark=None] table[x=factor,y=sdgcnnbody, col sep=comma] {iou2.csv};
		\end{axis}
	\end{tikzpicture}\hspace{0.01\textwidth}\begin{tikzpicture}
		\begin{axis}[
			title={Finger joints},
			xlabel={Distance threshold factor},
			ylabel={Percentage of Correct Joints},
			width=0.38\textwidth,
			every axis plot/.append style={very thick},
			grid style=dashed,
			ytick={0, 0.1, 0.2, 0.3, 0.4, 0.5, 0.6, 0.7, 0.8, 0.9, 1.0},
			yticklabel={\pgfmathparse{\tick*100}\pgfmathprintnumber{\pgfmathresult}\%},
			ymax=1.05,
			ymajorgrids=true,
			xmajorgrids=true,
			every tick label/.append style={font=\footnotesize},
			legend pos=outer north east
			]
			\addplot+[red, mark=None] table[x=factor,y=tarighands, col sep=comma] {iou2.csv};
			\addplot+[orange, mark=None] table[x=factor,y=sdgcnnhandsnn, col sep=comma] {iou2.csv};
			\addplot+[green, mark=None] table[x=factor,y=sdgcnnhands, col sep=comma] {iou2.csv};
			\legend{TARig--TJM, Ours (no normals), Ours (with normals)}
		\end{axis}
	\end{tikzpicture}
	\caption{Results of evaluation on the test set by the PCJ metric. Our method achieves significant improvement in both body and finger joint localization.}
	\label{figure:pck}
\end{figure*}
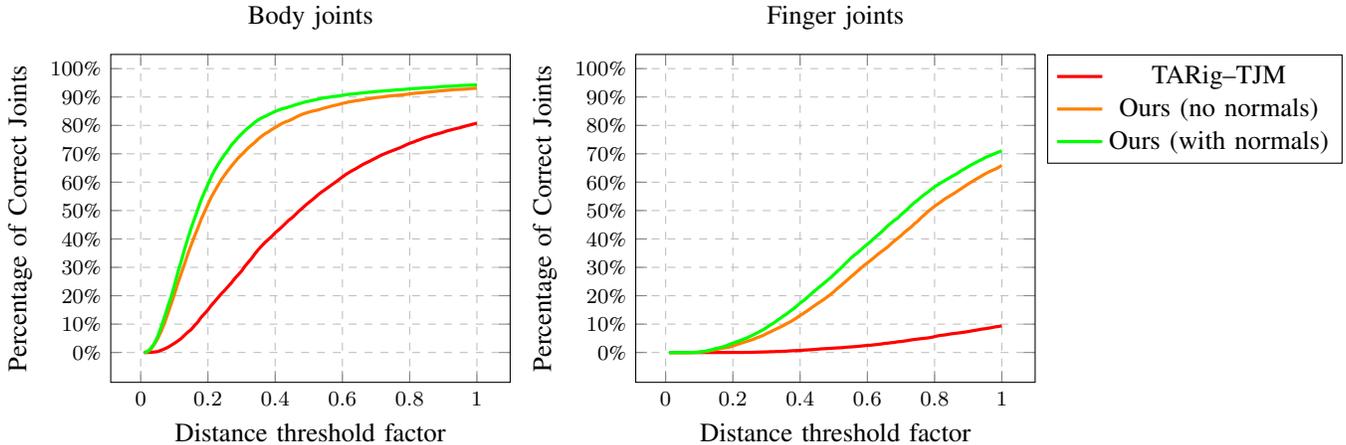

\subsection{Pose randomization}
The generated models are initially positioned in the A-pose. Since we want to evaluate the robustness of our neural network to pose variation-induced mesh deformation, we modify the mesh pose using the animation skeleton by arbitrarily rotating bones within predefined ranges up to 15 degrees per axis, using the Blender scripting API. The bones are rotated in a random order every time, with candidate orientations being tested for collisions before acceptance. Due to the complexity of collision checking, we employ an efficient raycast based algorithm. After a rotation has been performed, we emit rays from all bone origins in bone directions and check whether the number of obstacles hit by the ray has changed. If so, the rotation is rejected and a new one is generated and tested.

\subsection{Remeshing and normalization}
To reduce computation time and resource requirements, we simplify the mesh using the MeshLab \cite{10.2312:LocalChapterEvents:ItalChap:ItalianChapConf2008:129-136} implementation of the quadric edge collapse algorithm described in \cite{10.1145/3596711.3596727}, reducing the amount of mesh faces to $20480$ and mesh vertices to between $10000$ and $12000$. In order to achieve scale invariance, we also normalize the meshes to a unit cube by scaling down vertex positions by the height of the mesh. Each mesh is translated so that the center of its bounding box is at $(0, 0, 0)$. We convert the mesh to point cloud form using the Trimesh library for 3D mesh processing \cite{trimesh}.

\section{Experimental results}

\subsection{Training}
Our training and evaluation code is based on the PointMeta \cite{10205335} code base which wraps the OpenPoints engine, written in PyTorch and published as part of PointNeXt \cite{qian2022pointnext}. All experiments were performed on a computer with a single NVIDIA RTX 4070 12GB VRAM GPU, AMD Ryzen 3600 CPU and 16 GB of RAM. We train our proposed neural network for $100$ epochs with a batch size of $1$, using the AdamW optimizer \cite{Loshchilov2017DecoupledWD} with weight decay set to $1\mathrm{e}{-4}$ and the ReduceLROnPlateau learning rate scheduler with the initial learning rate set to $1\mathrm{e}{-3}$, patience set to $8$ epochs, decay rate set to $0.75$ and number of warmup epochs set to $0$. The neural network, given only a set of 3D point coordinates and their normal vectors, predicts the 3D locations of 29 body joints and 40 finger joints. For estimating normals of points in a point set, we use the Open3D library for 3D data processing \cite{Zhou2018}. During computation, the DGCNN EdgeConv modules compute $k{=}80$ neighbours using the $k$-nearest neighbours algorithm. The first EdgeConv layer computes neighbourhoods based on Euclidean distances in 3D space, while subsequent EdgeConv stages infer neighbourhoods based on Euclidean distances in feature space. During training, the data was augmented with point cloud scaling in the $[0.8, 1.2]$ range and point cloud jitter with $\sigma{=}0.01$ and $0.05$ clipping threshold.

\subsection{Evaluation}
We use two evaluation metrics: 1) Mean Per Joint Position Error (MPJPE) and 2) Percentage of Correct Joints (PCJ). Since the source code for TARig \cite{MA2023158} is not published, we reimplement their template joint module (TARig--TJM) for comparison purposes. Due to the observed architectural similarities, we base our reimplementation on the publicly available RigNet \cite{RigNet} code base.

The results of the method comparison using the MPJPE metric are shown in Table \ref{table:results_mae}. The errors are measured after the normalization to a unit cube and represent percentages of the 3D model height. For each joint category, our trained network produces smaller error values compared to TARig--TJM, even when normal vector estimates are omitted. We note that TARig--TJM processes the normal vectors.

To consider a joint correctly localized for the PCJ metric, the distance between the predicted joint and the groundtruth joint must be within a precomputed threshold. To compute this threshold for an individual joint, rays are emitted, for each bone containing that joint, from the groundtruth joint through the plane perpendicular to that bone, in a circular manner. If all rays miss the mesh, we repeat this action with the ray origin being moved towards the parent joint. When all ray-mesh intersection points are computed, we calculate the average distance between the groundtruth joint and the closest half of the ray-mesh intersection points. This distance is then multiplied by a factor between $0.01$ and $1.00$. The resulting value is the acceptance threshold for the distance between the predicted and the groundtruth joint. In Fig.~\ref{figure:pck} the PCJ measurements are shown as a function of the distance threshold factor. For all threshold values, our solution shows significantly better results.

\begin{figure*}
	\centering
	\subfloat[]{
		\adjincludegraphics[width=0.3\textwidth, trim={{.297\width} 0 {.258\width} 0}]{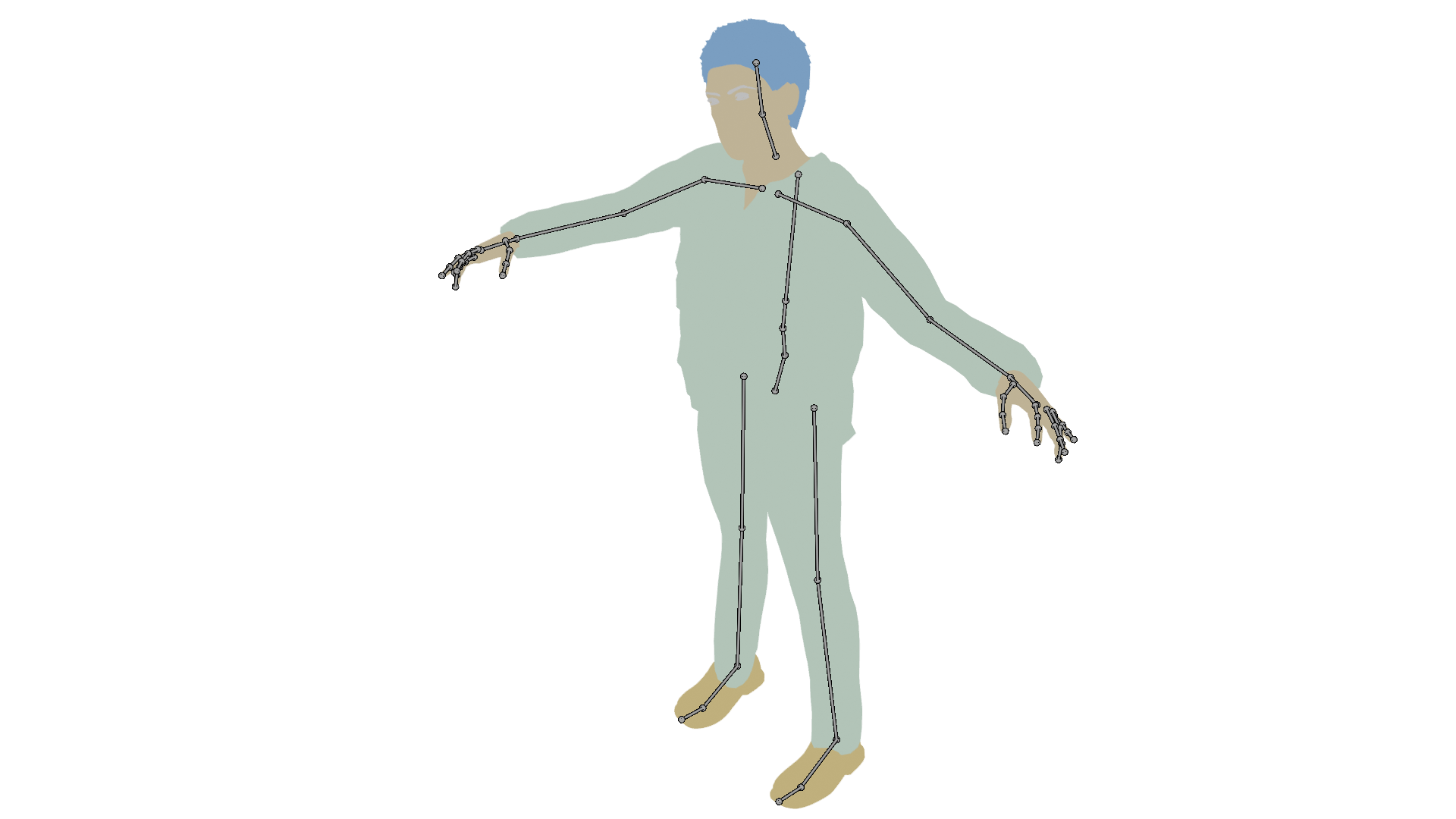}
		\label{figure:body_groundtruth}}
	\hfil
	\subfloat[]{
		\adjincludegraphics[width=0.3\textwidth, trim={{.297\width} 0 {.258\width} 0}]{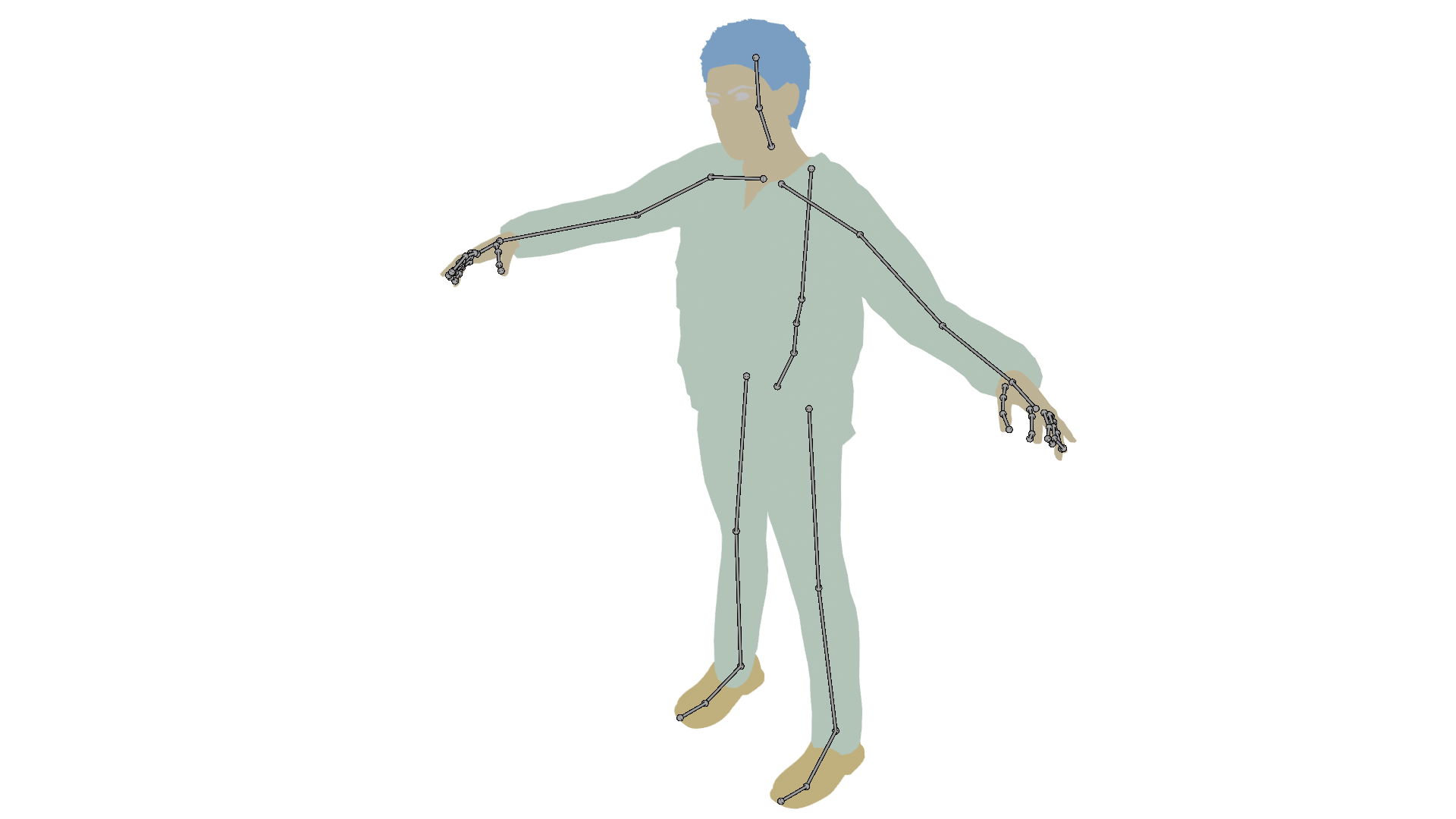}
		\label{figure:body_tarig}}
	\hfil
	\subfloat[]{
		\adjincludegraphics[width=0.3\textwidth, trim={{.297\width} 0 {.258\width} 0}]{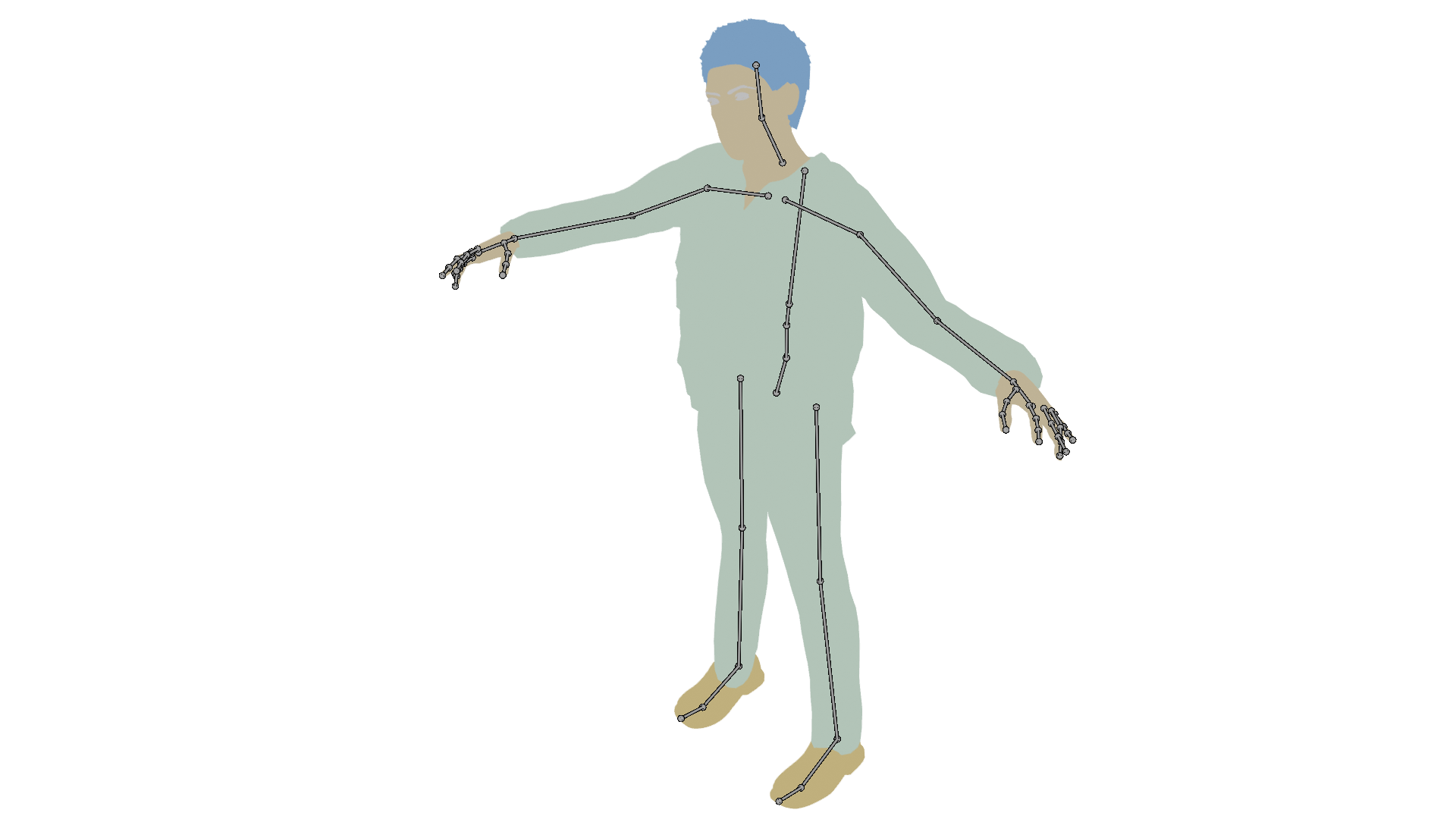}
		\label{figure:body_predictions_ours}}
	\caption{Comparison of body joint visualizations, rendered using Blender viewport rendering: (a) groundtruth, (b) TARig--TJM and (c) our method. Our method shows slight improvements which are noticeable on the joints of the spine and the right knee joint.}
	\label{figure:body_predictions}
\end{figure*}

\begin{figure*}
	\centering
	\subfloat[]{
		\adjincludegraphics[width=0.3\textwidth, trim={{.0\width} 0 {.145833\width} 0}]{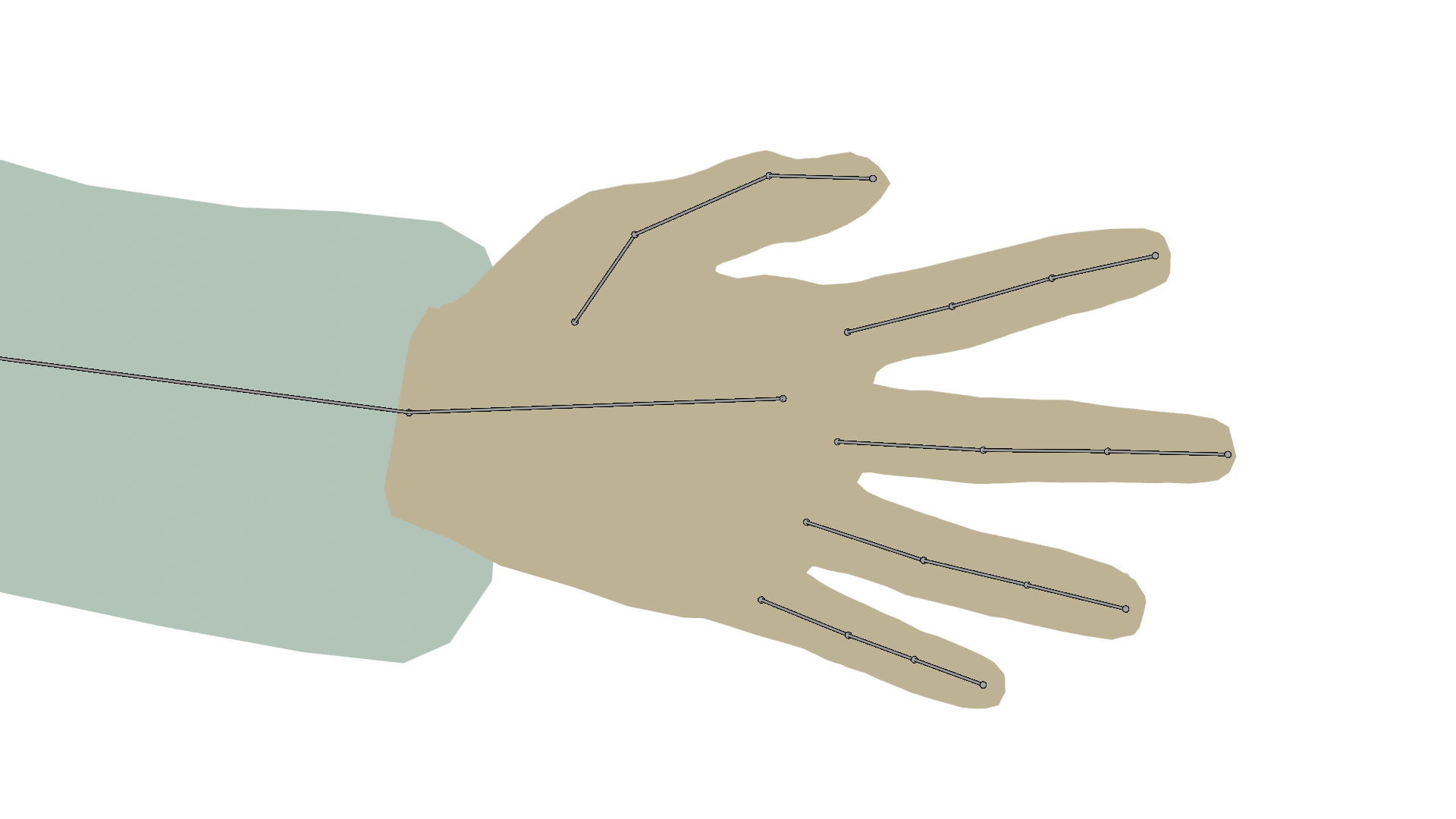}
		\label{figure:hand_groundtruth}}
	\hfil
	\subfloat[]{
		\adjincludegraphics[width=0.3\textwidth, trim={{.0\width} 0 {.145833\width} 0}]{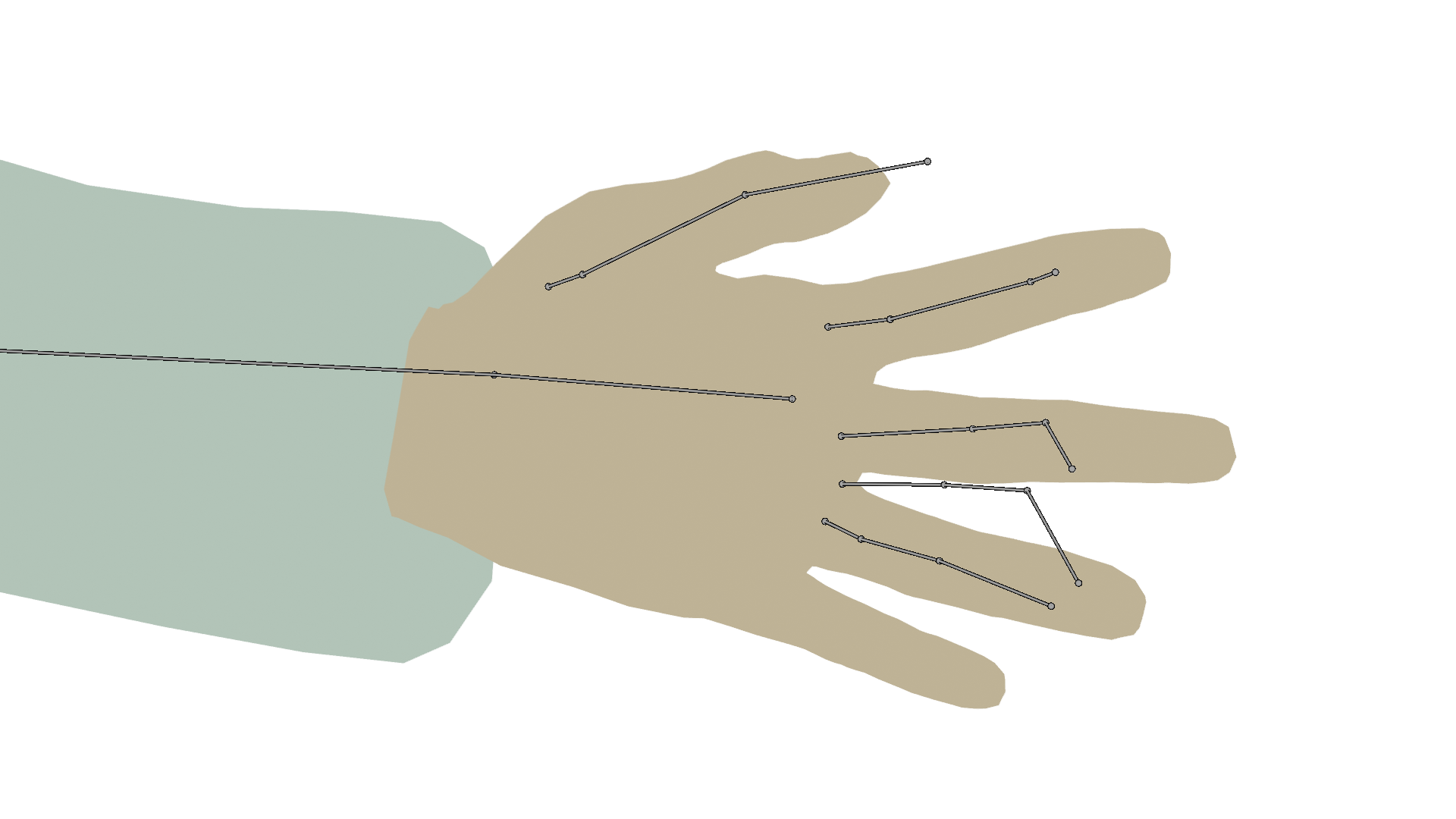}
		\label{figure:hand_tarig}}
	\hfil
	\subfloat[]{
		\adjincludegraphics[width=0.3\textwidth, trim={{.0\width} 0 {.145833\width} 0}]{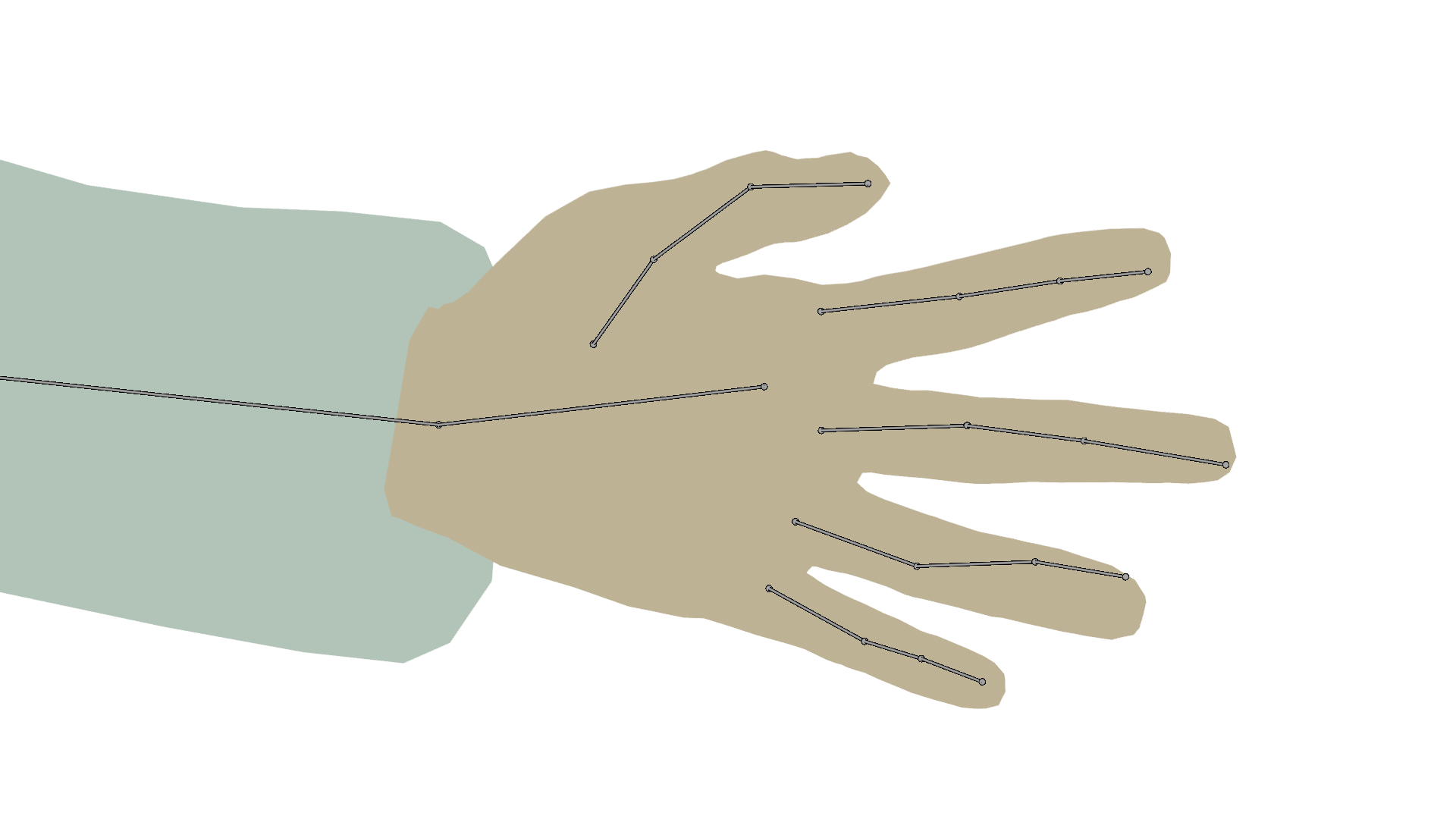}
		\label{figure:hand_predictions_ours}}
	\caption{Comparison of finger joint visualizations, rendered using Blender viewport rendering: (a) groundtruth, (b) TARig--TJM estimations and (c) estimations by our method, which show considerable improvements compared to TARig--TJM estimations. TARig--TJM places the little finger joints the ring finger, while the ring finger joints are outside of the mesh. The estimations by our method are devoid of such significant issues.}
	\label{figure:hand_predictions}
\end{figure*}

Body and finger joint estimation examples are shown in Fig.~\ref{figure:body_predictions} and Fig.~\ref{figure:hand_predictions}, respectively.
Our method shows improvements compared to TARig--TJM, especially for finger joints.

Our solution processes a 3D model in \qty{1.5}{\second}, while our TARig--TJM implementation requires \qty{25}{\second}. The latter is due to precomputation of geodesic neighbourhoods and attention maps, which take \qty{18}{\second} and \qty{6}{\second}, respectively.
Our network has $0.39M$ parameters. TARig--TJM has $1.78M$ parameters, most of which are shared with other TARig modules, such as those for boneflow and skinning, in the same backbone network.

\section{Conclusion and further work}
We have achieved state-of-the-art results in localizing body and finger joints for the animation skeleton of 3D human models, providing an efficient solution that necessitates only minimal data preprocessing. Our suggested method of generating synthetic data with varying shapes and poses can be used in further research to compare joint localization solutions. The proposed solution can be adapted for other datasets, including actual 3D scans, which makes it useful for reducing manual labor required when rigging 3D scans of humans. Our research code is published online \cite{ourresearchcode}.

Regarding future work, we suggest:
\begin{enumerate}
	\item Improving finger joint localization with a separate dedicated neural network.
	\item Relying on the knowledge transfer principle, pretraining a neural network using synthetic data before fine-tuning it on a dataset of scanned people.
	\item Applying pose randomization as a form of data augmentation during the training process.
\end{enumerate}

\bibliography{bib}

\begin{thebibliography}{10}

\bibitem{SMPL:2015}
M.~Loper, N.~Mahmood, J.~Romero, G.~Pons-Moll, and M.~J. Black, ``{SMPL}: A
  skinned multi-person linear model,'' {\em {ACM Transactions on Graphics,
  (Proc. SIGGRAPH Asia)}}, vol.~34, pp.~248:1--248:16, Oct. 2015.

\bibitem{MANO:SIGGRAPHASIA:2017}
J.~Romero, D.~Tzionas, and M.~J. Black, ``Embodied hands: Modeling and
  capturing hands and bodies together,'' {\em ACM Transactions on Graphics,
  (Proc. SIGGRAPH Asia)}, vol.~36, Nov. 2017.

\bibitem{SMPL-X:2019}
G.~Pavlakos, V.~Choutas, N.~Ghorbani, T.~Bolkart, A.~A.~A. Osman, D.~Tzionas,
  and M.~J. Black, ``Expressive body capture: {3D} hands, face, and body from a
  single image,'' in {\em IEEE/CVF Conference on Computer Vision and Pattern
  Recognition (CVPR)}, pp.~10975--10985, 2019.

\bibitem{STAR:2020}
A.~A.~A. Osman, T.~Bolkart, and M.~J. Black, ``{STAR}: A sparse trained
  articulated human body regressor,'' in {\em European Conference on Computer
  Vision (ECCV)}, pp.~598--613, 2020.

\bibitem{50649}
H.~Xu, E.~G. Bazavan, A.~Zanfir, B.~Freeman, R.~Sukthankar, and
  C.~Sminchisescu, ``{GHUM} \& {GHUML}: Generative {3D} human shape and
  articulated pose models,'' in {\em IEEE/CVF Conference on Computer Vision and
  Pattern Recognition (CVPR)}, pp.~6184--6193, 2020.

\bibitem{SUPR:2022}
A.~A.~A. Osman, T.~Bolkart, D.~Tzionas, and M.~J. Black, ``{SUPR}: A sparse
  unified part-based human body model,'' in {\em European Conference on
  Computer Vision (ECCV)}, 2022.

\bibitem{MakeHuman}
``{MakeHuman}.'' Available
  \url{https://github.com/makehumancommunity/makehuman}.
\newblock [Accessed: 19 February 2024].

\bibitem{9603558}
Z.~Fang, L.~Cai, and G.~Wang, ``{MetaHuman Creator}: The starting point of the
  metaverse,'' in {\em 2021 International Symposium on Computer Technology and
  Information Science (ISCTIS)}, pp.~154--157, 2021.

\bibitem{805368}
K.~Robinette, H.~Daanen, and E.~Paquet, ``The {CAESAR} project: a {3-D} surface
  anthropometry survey,'' in {\em Second International Conference on 3-D
  Digital Imaging and Modeling (Cat. No.PR00062)}, pp.~380--386, 1999.

\bibitem{6682899}
C.~Ionescu, D.~Papava, V.~Olaru, and C.~Sminchisescu, ``{Human3.6M}: Large
  scale datasets and predictive methods for {3D} human sensing in natural
  environments,'' {\em IEEE Transactions on Pattern Analysis and Machine
  Intelligence}, vol.~36, no.~7, pp.~1325--1339, 2014.

\bibitem{6909880}
F.~Bogo, J.~Romero, M.~Loper, and M.~J. Black, ``{FAUST}: Dataset and
  evaluation for {3D} mesh registration,'' in {\em IEEE/CVF Conference on
  Computer Vision and Pattern Recognition (CVPR)}, pp.~3794--3801, 2014.

\bibitem{zchao}
C.~Zhang, S.~Pujades, M.~Black, and G.~Pons-Moll, ``Detailed, accurate, human
  shape estimation from clothed {3D} scan sequences,'' in {\em IEEE/CVF
  Conference on Computer Vision and Pattern Recognition (CVPR)}, (Los Alamitos,
  CA, USA), pp.~5484--5493, IEEE Computer Society, July 2017.

\bibitem{8491001}
A.~Saint, E.~Ahmed, A.~E.~R. Shabayek, K.~Cherenkova, G.~Gusev, D.~Aouada, and
  B.~Ottersten, ``{3DBodyTex}: Textured {3D} body dataset,'' in {\em
  International Conference on 3D Vision (3DV)}, pp.~495--504, 2018.

\bibitem{yan2019anthropometric}
S.~Yan, J.~Wirta, and J.-K. Kämäräinen, ``Anthropometric clothing
  measurements from {3D} body scans,'' {\em Machine Vision and Applications},
  vol.~31, 2019.

\bibitem{Zheng2019DeepHuman}
Z.~Zheng, T.~Yu, Y.~Wei, Q.~Dai, and Y.~Liu, ``{DeepHuman}: {3D} human
  reconstruction from a single image,'' in {\em The IEEE International
  Conference on Computer Vision (ICCV)}, Oct. 2019.

\bibitem{tiwari20sizer}
G.~Tiwari, B.~L. Bhatnagar, T.~Tung, and G.~Pons-Moll, ``{SIZER}: A dataset and
  model for parsing {3D} clothing and learning size sensitive {3D} clothing,''
  in {\em European Conference on Computer Vision ({ECCV})}, {Springer}, Aug.
  2020.

\bibitem{tao2021function4d}
T.~Yu, Z.~Zheng, K.~Guo, P.~Liu, Q.~Dai, and Y.~Liu, ``{Function4D}: Real-time
  human volumetric capture from very sparse consumer {RGBD} sensors,'' in {\em
  IEEE/CVF Conference on Computer Vision and Pattern Recognition (CVPR)}, June
  2021.

\bibitem{Jinka2022SHARPSR}
S.~S. Jinka, R.~Chacko, A.~Srivastava, A.~Sharma, and P.~J. Narayanan,
  ``{SHARP}: Shape-aware reconstruction of people in loose clothing,'' {\em
  International Journal of Computer Vision}, vol.~131, pp.~918--937, 2022.

\bibitem{10125586}
X.~Zhu, T.~Liao, X.~Zhang, J.~Lyu, Z.~Chen, Y.~Wang, K.~Guo, Q.~Cao, S.~Z. Li,
  and Z.~Lei, ``{MVP-Human Dataset} for {3-D} clothed human avatar
  reconstruction from multiple frames,'' {\em IEEE Transactions on Biometrics,
  Behavior, and Identity Science}, vol.~5, no.~4, pp.~464--475, 2023.

\bibitem{9760157}
Z.~Su, T.~Yu, Y.~Wang, and Y.~Liu, ``{DeepCloth}: Neural garment representation
  for shape and style editing,'' {\em IEEE Transactions on Pattern Analysis and
  Machine Intelligence}, vol.~45, no.~2, pp.~1581--1593, 2023.

\bibitem{Blender}
``{Blender}.'' Available \url{https://www.blender.org}.
\newblock [Accessed: 20 February 2024].

\bibitem{ISOIEC19774-1:2019}
``{ISO/IEC 19774-1:2019} {Humanoid} animation ({H-Anim}) -- {Part 1}:
  Architecture.'' International Organization for Standardization, 2019.
\newblock Available from ISO (https://www.iso.org/standard/74163.html).

\bibitem{ISOIEC19774-2:2019}
``{ISO/IEC 19774-2:2019} {Humanoid} animation ({H-Anim}) -- {Part 2}: Motion
  capture data.'' International Organization for Standardization, 2019.
\newblock Available from ISO (https://www.iso.org/standard/74164.html).

\bibitem{10.5555/2385444}
R.~Parent, {\em Computer Animation: Algorithms and Techniques}.
\newblock San Francisco, CA, USA: Morgan Kaufmann Publishers Inc., 3~ed., 2012.

\bibitem{bhatnagar2020ipnet}
B.~L. Bhatnagar, C.~Sminchisescu, C.~Theobalt, and G.~Pons-Moll, ``Combining
  implicit function learning and parametric models for {3D} human
  reconstruction,'' in {\em European Conference on Computer Vision ({ECCV})},
  {Springer}, Aug. 2020.

\bibitem{bhatnagar2020loopreg}
B.~L. Bhatnagar, C.~Sminchisescu, C.~Theobalt, and G.~Pons-Moll, ``{LoopReg}:
  Self-supervised learning of implicit surface correspondences, pose and shape
  for {3D} human mesh registration,'' in {\em Neural Information Processing
  Systems (NeurIPS)}, Dec. 2020.

\bibitem{RigNet}
Z.~Xu, Y.~Zhou, E.~Kalogerakis, C.~Landreth, and K.~Singh, ``{RigNet}: Neural
  rigging for articulated characters,'' {\em ACM Transactions on Graphics},
  vol.~39, pp.~58:1--58:14, Aug. 2020.

\bibitem{MA2023158}
J.~Ma and D.~Zhang, ``{TARig}: Adaptive template-aware neural rigging for
  humanoid characters,'' {\em Computers \& Graphics}, vol.~114, pp.~158--167,
  2023.

\bibitem{8099499}
R.~Q. Charles, H.~Su, M.~Kaichun, and L.~J. Guibas, ``{PointNet}: Deep learning
  on point sets for {3D} classification and segmentation,'' in {\em IEEE/CVF
  Conference on Computer Vision and Pattern Recognition (CVPR)}, pp.~77--85,
  2017.

\bibitem{dgcnn}
Y.~Wang, Y.~Sun, Z.~Liu, S.~E. Sarma, M.~M. Bronstein, and J.~M. Solomon,
  ``Dynamic graph {CNN} for learning on point clouds,'' {\em ACM Transactions
  on Graphics (TOG)}, 2019.

\bibitem{qian2022pointnext}
G.~Qian, Y.~Li, H.~Peng, J.~Mai, H.~Hammoud, M.~Elhoseiny, and B.~Ghanem,
  ``{PointNeXt}: Revisiting {PointNet++} with improved training and scaling
  strategies,'' in {\em Advances in Neural Information Processing Systems
  (NeurIPS)}, 2022.

\bibitem{10205335}
H.~Lin, X.~Zheng, L.~Li, F.~Chao, S.~Wang, Y.~Wang, Y.~Tian, and R.~Ji, ``Meta
  architecture for point cloud analysis,'' in {\em IEEE/CVF Conference on
  Computer Vision and Pattern Recognition (CVPR)}, (Los Alamitos, CA, USA),
  pp.~17682--17691, IEEE Computer Society, June 2023.

\bibitem{lt3dhpfpc}
Y.~Zhou, H.~Dong, and A.~El~Saddik, ``Learning to estimate {3D} human pose from
  point cloud,'' {\em IEEE Sensors Journal}, vol.~PP, pp.~1--1, June 2020.

\bibitem{400568}
Y.~Cheng, ``Mean shift, mode seeking, and clustering,'' {\em IEEE Transactions
  on Pattern Analysis and Machine Intelligence}, vol.~17, no.~8, pp.~790--799,
  1995.

\bibitem{yin2018p2pnet}
K.~Yin, H.~Huang, D.~Cohen-Or, and H.~Zhang, ``{P2P-NET}: Bidirectional point
  displacement net for shape transform,'' {\em ACM Transactions on Graphics
  (Special Issue of SIGGRAPH)}, vol.~37, no.~4, pp.~152:1--152:13, 2018.

\bibitem{Rockafellar+1970}
R.~T. Rockafellar, {\em Convex Analysis}.
\newblock Princeton: Princeton University Press, 1970.

\bibitem{10.2312:LocalChapterEvents:ItalChap:ItalianChapConf2008:129-136}
P.~Cignoni, M.~Callieri, M.~Corsini, M.~Dellepiane, F.~Ganovelli, and
  G.~Ranzuglia, ``{MeshLab}: An open-source mesh processing tool,'' in {\em
  Eurographics Italian Chapter Conference} (V.~Scarano, R.~D. Chiara, and
  U.~Erra, eds.), The Eurographics Association, 2008.

\bibitem{10.1145/3596711.3596727}
M.~Garland and P.~S. Heckbert, {\em Surface Simplification Using Quadric Error
  Metrics}.
\newblock New York, NY, USA: Association for Computing Machinery, 1~ed., 2023.

\bibitem{trimesh}
``{Trimesh}.'' Available \url{https://trimesh.org/}.
\newblock [Accessed: 20 February 2024].

\bibitem{Loshchilov2017DecoupledWD}
I.~Loshchilov and F.~Hutter, ``Decoupled weight decay regularization,'' in {\em
  International Conference on Learning Representations}, 2017.

\bibitem{Zhou2018}
Q.-Y. Zhou, J.~Park, and V.~Koltun, ``{Open3D}: {A} modern library for {3D}
  data processing,'' {\em arXiv:1801.09847}, 2018.

\bibitem{ourresearchcode}
``Learning localization of body and finger animation skeleton joints on
  three-dimensional models of human bodies).'' Available
  \url{https://github.com/sznov/joint-localization}.
\newblock [Accessed: 27 February 2024].

\end{thebibliography}
\bibliographystyle{ieeetr}

\end{document}